\batchmode
\makeatletter
\def\input@path{{/home/rilab3/Desktop/Bpaper/RoboticGraspfromNLDescription/}}
\makeatother
\documentclass[english]{IEEEtran}
\usepackage[T1]{fontenc}
\usepackage{array}
\usepackage{float}
\usepackage{booktabs}
\usepackage{amsmath}
\usepackage{amssymb}
\usepackage{graphicx}
\usepackage{rotating}

\makeatletter

\newcommand{\lyxmathsym}[1]{\ifmmode\begingroup\def\b@ld{bold}
  \text{\ifx\math@version\b@ld\bfseries\fi#1}\endgroup\else#1\fi}

\providecommand{\tabularnewline}{\\}
\floatstyle{ruled}
\newfloat{algorithm}{tbp}{loa}
\providecommand{\algorithmname}{Algorithm}
\floatname{algorithm}{\protect\algorithmname}

\usepackage{cite}
\usepackage{colortbl}
\definecolor{lightgray}{gray}{0.95}

\AtBeginDocument{
  
}

\makeatother

\usepackage{babel}
\begin{document}
\title{Knowledge-Augmented Dexterous Grasping with Incomplete Sensing}
\author{Bharath Rao, Hui Li, Krishna Krishnan, Enkhsaikhan Boldsaikhan, and
Hongsheng He$^{*}$\textit{ }\thanks{Bharath Rao is with Cognitive Robotics, Spirit AeroSystems, Wichita,
KS, 67260, USA }\thanks{Hui Li and Hongsheng He are with the Department of Electrical Engineering
and Computer Science, Wichita State University, Wichita, KS, 67260,
USA}\thanks{Krishna Krishnan is with the Department of Industrial, Systems, and
Manufacturing Engineering, Wichita State University, Wichita, KS,
67260, USA}\thanks{Enkhsaikhan Boldsaikhan is with the Department of Industrial, Systems,
and Manufacturing Engineering, Wichita State University, Wichita,
KS, 67260, USA }\thanks{{*}Correspondence should be addressed to Hongsheng He, hongsheng.he@wichita.edu.}}
\maketitle
\begin{abstract}
Humans can determine a proper strategy to grasp an object according
to the measured physical attributes or the prior knowledge of the
object. This paper proposes an approach to determining the strategy
of dexterous grasping by using an anthropomorphic robotic hand simply
based on a label or a description of an object. Object attributes
are parsed from natural-language descriptions and augmented with an
object knowledge base that is scraped from retailer websites. A novel
metric named joint probability distance is defined to measure distance
between object attributes. The probability distribution of grasp types
for the given object is learned using a deep neural network which
takes in object features as input. The action of the multi-fingered
hand with redundant degrees of freedom (DoF) is controlled by a linear
inverse-kinematics model of grasp topology and scales. The grasping
strategy generated by the proposed approach is evaluated both by simulation
and execution on a Sawyer robot with an AR10 robotic hand.
\end{abstract}


\section{Introduction}

Dexterous grasping is critical for complex assembly and delicate tool
handling in industrial automation and advanced manufacturing \cite{li2016dexterous}.
Dexterous robotic manipulation replies on comprehensive and precise
measurement of the work context \cite{hui20magic}, which is usually
impractical and expensive for industrial applications. It is even
challenging to measure important parameters of objects for grasping
in an online mode, such as 3D dimensions, materials, and weights.
It is therefore beneficial to research an approach to plan dexterous
grasping without complete or accurate sensing of object characteristics. 

A five-digit hand configuration with an opposable thumb is considered
to be one of the most important natural selection that contributed
to human evolutionary success. There are considerable application
scenarios where a dexterous robotic hand could be invaluable such
as disaster struck areas where the robot may have to interact with
unfamiliar environment. It has been an elusive goal to enable a robot
to master the human-level grasping skills. Recent studies on robotic
grasping have focused on two or three-fingered robotic clamps \cite{PelossofSVMGrasp,Spiers2Finger}.
The research of grasping planning for an anthropomorphic robotic hand
is more challenging and deserves more effort \cite{BonilaSoftHands5Fing,Morales2006VisualPlanning5fingers}. 

Behind this simple task of grasping an object, the human brain is
executing a series of sub-tasks with the associated decision-making
and error-correction process in real time. The brain selects and executes
appropriate motor strategy learned earlier by the human sensorimotor
apparatus. These learned strategies, also called as action-phase controllers
\cite{NatureMechanoreceptors}, utilize the input sensory signals
and corresponding predictions by the nervous system to produce motor
commands to accomplish the given motor task. The action-phase controllers
accurately estimate the specific motor output required using the information
about an object\textquoteright s physical properties and the current
configuration of the hand. The inspiration for this study is the human
grasping mechanism - not only from the sensorimotor task of human
grasping, but also the learning process itself. This study is therefore,
an exploration in emulating a part of the human brain\textquoteright s
action-phase controller model to accomplish grasping using a humanoid
robotic hand.

\begin{figure}
\includegraphics[width=1\columnwidth]{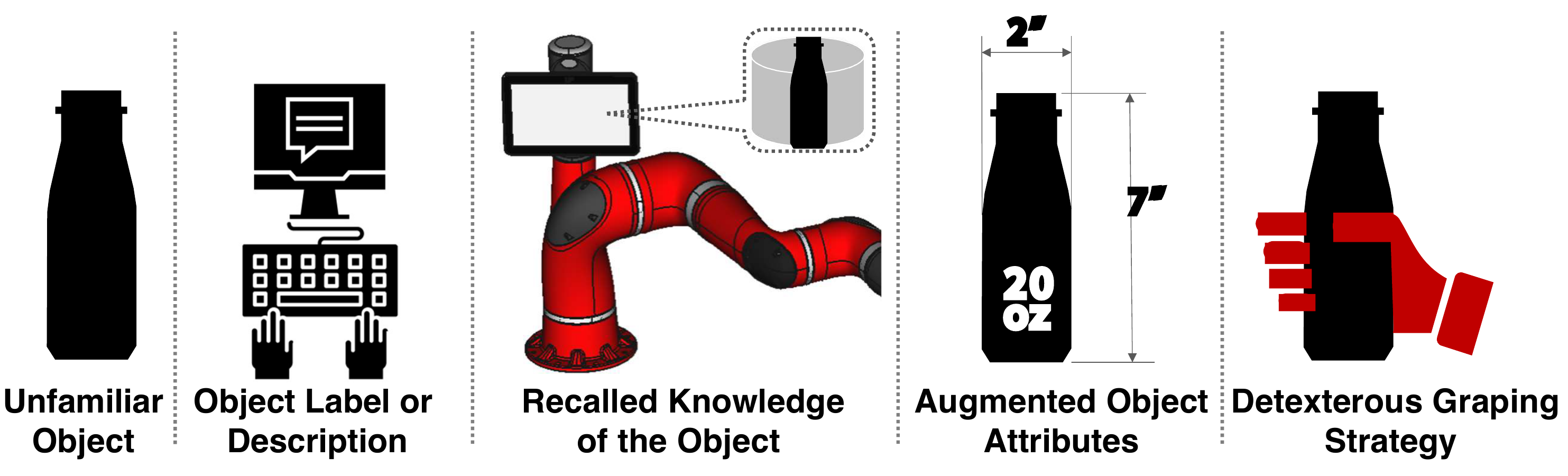}\caption{Framework of the proposed approach.\label{fig:Framework-of-the}}
\end{figure}

In this paper, we propose an approach to emulating human grasping
strategies without complete sensing of object attributes, as shown
in Fig.~\ref{fig:Framework-of-the}. By using labels and descriptions,
object attributes are retrieved and augmented from a knowledge base
that is scraped from online webpages. The extended object attributes
include dimension, mass, shape, texture, fragility, material, and
stiffness. We design a neural-network model to learn human grasping
strategies for target objects with various physical attributes. The
optimal grasping strategy is deployed to the anthropomorphic robot
hand by a multi-constrained inverse kinematics of grasping topology
and scales.

Human hands have 20 degrees-of-freedom (DoF) each (not including the
wrist joint), thousands of mechanoreceptors \cite{NatureMechanoreceptors}
and therefore, a significant amount of the brain\textquoteright s
resources are dedicated to grasping tasks. Understanding human grasps
is not a trivial task. Most of the efforts in understanding grasps
\cite{Cutkosky1989,Feix2016l} have been to breakdown the human grasping
behavior into discrete classes. A structured classification of grasps
is discussed in \cite{Cutkosky1989} based on object shapes and task
requirements. More recently a new and more comprehensive version of
the Grasp taxonomy has been developed \cite{Feix2016l,Heinemann2015humangrasptransfertorobothands}
and refined by de-coupling them from the object shapes and the tasks
being performed. A neuroscience-based study reports that hand posture
can be decomposed into very few general configurations and that the
finer adjustments can be achieved by superposition of such grasp poses
\cite{SantelloNeurosciencePCA}. Built on this concept, a method of
using ``eigengrasps'' to reduce dimensionality of grasps was proposed
\cite{CiocarlieEigenGrasps}. Reducing dimensionality is a necessary
step to make the problem of learning grasps tractable. 

Robotic dexterity has long been a difficult goal. Earlier methods
involved analytical approaches to calculate object affordances and
contact forces to determine grasp successes \cite{Miller2003ShapePremitives,JainGraspDetect2fing}.
Knowledge based systems and expert systems have been employed to choose
grasps \cite{Cutkosky1989,Stransfield1991KBGrasping} where in the
mechanism of grasping are broken down into discrete deterministic
rules. But the sheer number of variations of human grasps and the
difficulty in modeling various grasp scenarios limit such approaches
to few narrow applications. Recent proposals have focused on learning
methods \cite{LiGraspUncertainity,Herzog2014Shape-templates,KROEMER2010GaussianBandits,AIVALIOTIS2017MLforCmpxPrts,MadryKragicPGM,Song10learningtask},
especially application of deep learning methods to learn grasps \cite{DBLP:journals/corr/PintoG15,LenzDeepLearn,SaxenaAndNGNovelObjectsVision}.

It is the redundant DoF of multi-fingered hands that enables dexterity
of grasping and manipulation. There may be many possible strategies
to grasp an object, and the optimal one depends on the affordances
of the target object. Humans have the ability to apply proper grasping
strategies for unfamiliar objects based on simple descriptions or
formed association with known similar objects. As complete sensing
of all object attributes is unworkable in industrial applications,
robots would stand a better chance of success if they can imitate
this human awareness and knowledge with incomplete sensing. We develop
a knowledge base by mining object attributes online, and identify
the best match from the dataset by using natural-language object descriptions
as input. 

The AR10 humanoid robotic hand in this work is is equipped with 10
servos and limited force feedback. Predicting 10 joint angles given
a list of object attributes, is an ill-posed problem due to the infinite
ways these joint angles can be configured. In this paper, a novel
method is being proposed to make this problem tractable. The joint
angle configuration space is discretized to consist of various human
grasp types, whereby only the specific grasp type needs to be learned.
The variations resulting out of differing object sizes, are addressed
by introducing a scaling factor to the joint angles. This approach
makes the problem of learning five-fingered grasps tractable by discretizing
the grasp space and reducing the dimensionality of the problem. The
problem of grasp selection could have been treated as a multi-class
classification problem involving the selection of one of the possible
grasps from the human grasp primitives. However, complexity exists
in grasp labeling. There is no one ideal way to grasp a given object.
Humans tend to choose grasps based on the object\textquoteright s
position, orientation, intended action and sometimes even making arbitrary
grasp choices. So, the problem is not to choose an ideal grasp, but
to choose one from a set of the feasible grasps, which perhaps would
also be a preferred human grasp. To achieve this, multiple human grasping
trials were conducted, and frequencies of the grasps were used as
the labels against each object. A deep neural network model was trained
on this labeled dataset to estimate probabilities of various grasp
types conditioned on object\textquoteright s physical features. The
success of the approach is evaluated by validating the most probable
(predicted) grasp against the feasible set of human labeled grasps.

Human grasping is a complex process with disproportionately large
portion of the human sensorimotor apparatus dedicated to grasping.
Therefore, it is no surprise that robotic grasping is a complex and
as yet unsolved problem. The contribution of this paper is to further
the knowledge and understanding of human grasping in the context of
emulating human type grasps on a five-fingered robotic hand. In order
to demonstrate this idea, a set of everyday objects are chosen to
train the robot, to impart the knowledge and experience that is needed
for it to succeed at grasping. Multiple learning models with novel
concepts are developed and validated in the course of exploring the
five-fingered robotic grasping problem. The models are tested through
simulations and experiments with the physical robot. The major contributions
of the paper include:
\begin{enumerate}
\item The paper addresses the challenge of acquiring object attributes without
complete sensing. A distance metric is proposed to query most similar
objects in the developed knowledge base by using simple object descriptions. 
\item The paper designs a neural-network model to imitate human abilities
in applying optimal grasping strategies for dexterous grasping. A
well designed grasping strategies with grasping topology allows dexterous
grasping of objects without precise attribute information. 
\end{enumerate}

\section{Object Affordance Acquisition from Knowledge Base}

It is generally challenging to measure complete object attributes
in industrial applications, including precise 3D dimension, materials,
rigidity, and textures, but object categories can be recognized by
machine learning algorithms. In addition, it is straightforward to
describe important object attributes in natural language. We therefore
developed a knowledge base of object attributes by mining online object
information from retailer websites. By referring to the knowledge
base, we can acquire extended object attributes by object labels or
short descriptions for the selection of optimal grasping strategies.
In this section, we define the dominant object attributes for dexterous
grasping, design the parsing algorithm for key attributes from natural-language
descriptions, and propose a novel distance metric for attribute acquisition
from the knowledge base. 

\subsection{Dominant Object Attributes in Grasping}

Human grasp strategies depend on numerous factors including object
shape, size, weight, texture, stiffness and sometimes fragility, temperature,
wetness \cite{NapierPrehenileMovements}. With our goal being able
to understand these object features by parsing natural-language descriptions,
we had to be parsimonious in our choice of object attributes. Based
on findings coming out of previous studies, the following set of features,
shown in Table \ref{tab:Description-of-Object} were prioritized for
data collection. These physical attributes of the object significantly
influence grasping decisions. 

\begin{table}[H]
\begin{centering}
\caption{Primary object attributes in grasping.\label{tab:Description-of-Object}}
\par\end{centering}
\centering{}%
\begin{tabular}[t]{>{\raggedright}p{0.1\columnwidth}>{\raggedright}p{0.35\columnwidth}>{\raggedright}p{0.4\columnwidth}}
\toprule 
\textbf{Feature} &
\textbf{Description} &
\textbf{Value Range}\tabularnewline
\midrule
\addlinespace[2pt]
\rowcolor{lightgray}$\left(a,b,c\right)$ &
Dimensions along orthogonal directions &
$\left(a,b,c\right)\in\mathbb{R}^{3}$ s.t. $a\geq b\geq c$\tabularnewline
\addlinespace[2pt]
$m$ &
Mass &
$m\in\mathbb{R}$\tabularnewline
\addlinespace[2pt]
\rowcolor{lightgray}$s$ &
Shape classification \cite{Cutkosky1989} &
thin, compact, prism, long, radial\tabularnewline
\addlinespace[2pt]
$r$ &
Rigidity of the object &
rigid, squeezable, floppy\tabularnewline
\addlinespace[2pt]
$t$ &
Texture &
medium, smooth, rough\tabularnewline
\addlinespace[2pt]
$fr$ &
Fragility &
sturdy, medium, fragile\tabularnewline
\addlinespace[2pt]
\rowcolor{lightgray}$mt$ &
Simplified material types &
fabric, glass, metal, paper, plastic, rubber, wood, other\tabularnewline
\bottomrule
\addlinespace[2pt]
\end{tabular}
\end{table}

\begin{table*}
\caption{Object knowledge base \label{tab:Object-Features-Dataset} (cf. Table
\ref{tab:Description-of-Object} for descriptions of the features).}

\centering{}%
\begin{tabular}{l>{\raggedright}p{0.2\textwidth}lllllllll}
\toprule 
\textbf{\#} &
\textbf{Object} &
\textbf{Length (a)} &
\textbf{Width (b)} &
\textbf{Height (c)} &
\textbf{Mass} &
\textbf{Shape} &
\textbf{Texture} &
\textbf{Fragility} &
\textbf{Material} &
\textbf{Stiffness}\tabularnewline
\midrule
\addlinespace[2pt]
\rowcolor{lightgray}1 &
calculator &
15.4 &
7.9 &
1.5 &
116 &
thin &
medium &
medium &
plastic &
rigid\tabularnewline
\addlinespace[2pt]
2 &
water bottle &
21.5 &
7.2 &
7.2 &
660 &
prism &
smooth &
sturdy &
metal &
rigid\tabularnewline
\addlinespace[2pt]
\rowcolor{lightgray}3 &
salt shaker &
8.2 &
3.1 &
3.1 &
82 &
prism &
smooth &
sturdy &
metal &
rigid\tabularnewline
\addlinespace[2pt]
4 &
computer mouse &
10.6 &
5.9 &
2.5 &
79 &
prism &
medium &
medium &
plastic &
rigid\tabularnewline
\addlinespace[2pt]
\rowcolor{lightgray}5 &
mini rubix cube &
3.0 &
3.0 &
3.0 &
12 &
compact &
smooth &
sturdy &
plastic &
rigid\tabularnewline
\addlinespace[2pt]
6 &
wood wedge &
6.0 &
3.0 &
1.5 &
11 &
prism &
rough &
sturdy &
wood &
rigid\tabularnewline
\addlinespace[2pt]
\rowcolor{lightgray}7 &
wood disk &
7.2 &
7.2 &
2.0 &
60 &
compact &
rough &
sturdy &
wood &
rigid\tabularnewline
\addlinespace[2pt]
8 &
tennis ball &
6.4 &
6.4 &
6.4 &
56 &
radial &
rough &
medium &
fabric &
soft\tabularnewline
\addlinespace[2pt]
\rowcolor{lightgray}9 &
stapler &
13.2 &
6.8 &
3.6 &
151 &
prism &
smooth &
sturdy &
plastic &
rigid\tabularnewline
\addlinespace[2pt]
10 &
kitchen scale &
20 &
20 &
1.8 &
830 &
thin &
smooth &
sturdy &
glass &
rigid\tabularnewline
\bottomrule
\addlinespace[2pt]
\end{tabular}
\end{table*}

Enormous object descriptions are available in the internet particularly
in retailer webpages such as Amazon and Walmart. The object descriptions
typically include object dimensions, weight, and materials. We developed
a web scraper to collect object information (the source code is available
online at https://github.com/hhelium). The web scraper downloads product
description webpages and uses pattern search to discover object attributes.
The knowledge base consisted of dimension measurements, mass, rigidity,
material, texture, fragility, and shape classifications. The missing
attributes for some objects are manually labeled and annotated. Examples
of the objects and attributes in the knowledge base are shown in Table.~\ref{tab:Object-Features-Dataset}.
Given object labels or short descriptions $l$ will be mapped to the
feature set $f$ corresponding to the described object or a similar
object
\begin{equation}
l\rightarrow f=[a,b,c,m,s,mt,r,t,fr]
\end{equation}
where the meanings of the attributes are defined in Table~\ref{tab:Description-of-Object}.
We will discuss the description parsing and mapping in the following
sections.  

\subsection{Parsing Object Descriptions \label{subsec:Translating-Object-Descriptions}}

The problem to address in parsing object descriptions is to estimate
significant object features that determine object categories. At the
same time, the method is required to be resilient to missing, partial
or incorrect descriptions. Object descriptions may specify object
details such as its approximate dimensions, e.g., \textquotedblleft it
is about ten centimeters long'', or materials, e.g., ``it is made
of plastic\textquotedblright . Though not accurate or specific, the
object descriptions are informative when the object's descriptive
and quantitative information is contained. To specifically address
free-form descriptions where object object features are described
in any format or order, we designed a natural-language parser as shown
in Algorithm \ref{alg:Natural-Language-Parsing}. The natural-language
statements are cleaned up by lemmatizing and removal of stop words.
Each word in the statements is then tagged with parts of speech (POS)
labels based on standardized tags from the Penn Treebank project \cite{PennTreebank}.
Examples of POS Tag Nomenclature include JJ-Adjective, IN-preposition
or conjunction, CD-Cardinal digit, CC-coordinating conjunction, NN-noun,
and RB-adverb. Of special interest to our study, are any available
quantitative and qualitative descriptors of the object(s). We look
for expressions such as \textquotedblleft two centimeters long\textquotedblright ,
\textquotedblleft made of plastic\textquotedblright{} or \textquotedblleft very
rough\textquotedblright{} using regular expressions as shown in the
algorithm. 

\begin{algorithm}[h]
Input: Object description string: ObjDescrption;

Input: Regular Expression: regex =

\{<JJ.?>{*}<IN>{*}<CD.?><CD.?>{*}

<CC.?>{*}<CD.?>{*}<NN.?>{*}<RB.?>

{*}<JJ.?>{*}<IN>{*}<NN.?>{*}<JJ.?>{*}<NN.?>?\}

Output: Array of dimensions of the object: $l=[a,b,c,m,...]$

\medskip{}

\textit{WordToks$\leftarrow$tokenize(ObjDescrption)}

\textit{POSToks$\leftarrow$ApplyPartsofSpeechTokens(WordToks)}

\textit{PhraseTree$\leftarrow$PerformChunking(POSToks, regex)}

\textbf{\textit{\emph{for}}}\textit{ each Chunk }\textbf{\textit{\emph{in}}}\textit{
PhraseTree}\textbf{\textit{ }}\textbf{\textit{\emph{do}}}

\textbf{\textit{\quad{}}}\textbf{\textit{\emph{if}}}\textit{ QuantitativeDescriptor
}\textbf{\textit{\emph{in}}}\textit{ Chunk}\textbf{\textit{ }}\textbf{\textit{\emph{then}}}

\textbf{\textit{\quad{}\quad{}}}\textit{$l_{i}$$\leftarrow$ParseToNumber(Chunk)}

\textbf{\textit{\quad{}}}\textbf{\textit{\emph{end if}}}

\textbf{\textit{\quad{}}}\textbf{\textit{\emph{if}}}\textit{ QualitativeDescriptor}\textbf{\textit{
}}\textbf{\textit{\emph{in}}}\textit{ Chunk}\textbf{\textit{ }}\textbf{\textit{\emph{then}}}

\textbf{\textit{\quad{}\quad{}}}\textit{$l_{i}$$\leftarrow$ParseToCategorical(Chunk)}

\textbf{\textit{\quad{}}}\textbf{\textit{\emph{end if}}}

\textbf{\textit{\quad{}}}\textbf{\textit{\emph{if}}}\textbf{\textit{
null }}\textbf{\textit{\emph{in}}}\textbf{\textit{ }}\textit{l$_{i}$
}\textbf{\textit{\emph{then}}}

\textbf{\textit{\quad{}\quad{}}}\textit{$l_{i}$$\leftarrow$DataImputation(l$_{i}$)}

\textbf{\textit{\quad{}}}\textbf{\textit{\emph{end if}}}

\textbf{\textit{\emph{end for}}}\caption{Object description parsing. \label{alg:Natural-Language-Parsing}}
\end{algorithm}

The extracted phrases (or chunks) lead us to individual feature descriptors.
When there are more than one dimensional descriptor, the largest
value is assigned to feature $a$, the smallest to feature $c$ and
the intermediate one to feature $b$. One of the challenges is that
if we do not have all feature descriptors then we have null values.
One example is when the object has a radial symmetry, it is described
only by diameter. For such cases, we perform data imputation using
a rule-based approach of estimating the missing dimension based on
the other available dimensions of the object. The rule itself was
derived from the priors in the data. The success of this model is
evaluated by scoring the parsed values with the measured or labeled
values and the scores are used to improve the algorithm.

\subsection{Object Knowledge Acquisition\label{subsec:Memory-recall-and}}

In addition to the basic object attributes in the description, we
desire to acquire more features from the knowledge base. Even with
a reasonably detailed elucidation of an object\textquoteright s features,
object descriptions tend to be either imprecise or incomplete. For
example, the general tendency is to round-off dimensions and mass,
often missing to mention certain features such as material type or
texture. A human can still work with the information available only
because of the recall of having seen or held such an object. The curated
knowledge base of objects and their physical features emulates human
memory. 

Let $f(o_{i})=[f_{1}^{i},f_{2}^{i},f_{3}^{i},\ldots,f_{m}^{i}]$ represent
$m$ features corresponding to an $i^{\textnormal{th}}$ object $o_{i}$
from this dataset where $i\text{\ensuremath{\in}}[1,N]$ and $f_{i}\in[a,b,c,m,s,mt,r,t,fr]$.
The features parsed from descriptions of a reference object $o_{r}$
is represented by $f(o)$. It should be noted that $f(o)$ could possibly
have empty values for some of the elements due to incomplete object
descriptions. The problem is to find the most similar object to the
reference object in the knowledge base $f^{*}=\arg\min_{i}\left\Vert o_{i}-o\right\Vert $.
By identifying an object closely matching the description of the reference
object, we continue to retain the ability of choosing the most suitable
grasp because the reference object, most likely, has physical features
very similar to the object chosen by the algorithm.

Several distance metrics, such as Euclidian, Minkowski and Cosine
distances, have been developed to  measure the proximity of vectors
in n-dimensional space \cite{ChaDistance,GomaaTextSimilarity}. To
increase the identification accuracy and the confidence of object
match, it is necessary to not only ensure the proximity of the points
in the normed vector space, but also to ensure proximity of each individual
feature. To that end, we calculate the probability of the $i^{\textnormal{th}}$
object being mapped to the reference object given the distance $|f_{j}(o_{i})-f_{j}(o)|$
of the $j^{\textnormal{th}}$ feature. The overall probability of
mapping $i^{\textnormal{th}}$ object to the reference object is the
joint probability over all the features of the object. This approach
ensures that only that object which matches the reference object\textquoteright s
every individual available features is the one that results in a highest
probability value. The proposed distance metric, named Joint Probability
Distance, is defined as
\begin{equation}
\left\Vert o_{i}-o_{r}\right\Vert =\sum_{j=1}^{m}\ln(1+|f_{j}(o_{i})-f_{j}(o_{r})|)
\end{equation}
One of the advantages of using this distance metric is that the probability
of the match decays at a much faster rate as each of the features
deviate from the reference. This helps in non-linearly increasing
the distance of unlikely candidates and filtering out the unlikely
matches with more confidence.

\begin{figure}
\includegraphics[width=1\columnwidth]{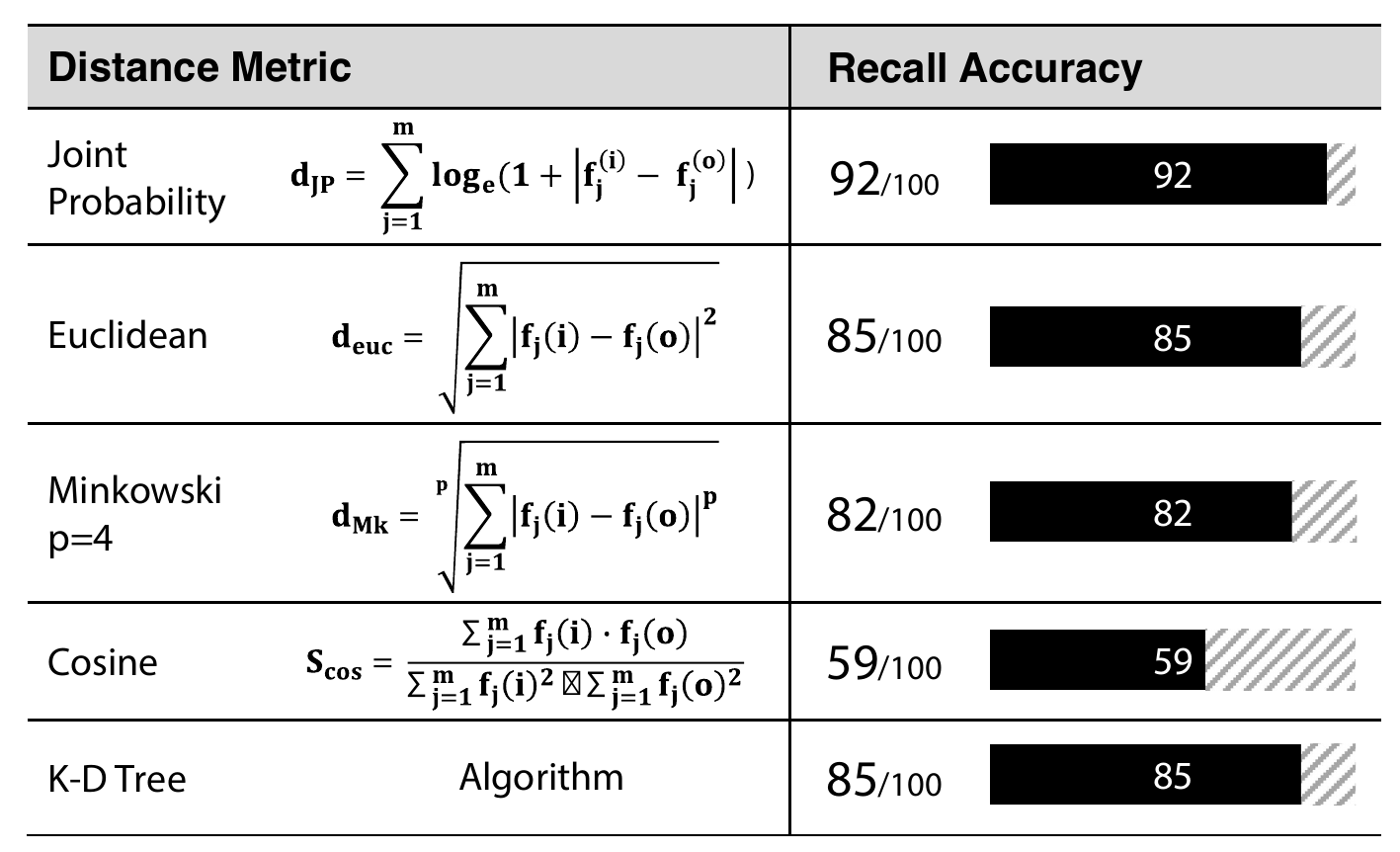}\caption{Comparison of object recall accuracy of various distance metrics.\label{fig:Comparison-of-object}}
\end{figure}

This distance metric can be used on a dataset with contains a combination
of continuous and categorical (transformed to one-hot binary encoding)
features without any need for data normalization. For example, material
of an object is a categorical feature with eight possible text values.
The Material property feature can be easily converted to eight features
with one-hot encoding. This distance metric will work well with such
categorical variables. The accuracy of distance metrics including
Euclidean, Minkowski, Cosine, K-D tree, and Joint Probability was
compared and results are shown in Fig.~\ref{fig:Comparison-of-object}.
The proposed Joint Probability metric achieved the best recall accuracy
for the knowledge base. 

\section{Human-Like Dexterous Grasping\label{subsec:Prediction---Learning}}

The grasp of a multi-finger robotic hand defines a set of angles of
the finger joints, and the magnitude of the contact forces applied
by the fingers and palm to an object at the contact points. The objective
here is to emulate human dexterous grasping by mapping object features
$f(o)$ referred in the knowledge base onto a grasp prioritization.
The referred object features may be inaccurate or even erroneous due
to rough measurement and description. We therefore implement grasp
strategies in terms of grasp topology and scales, which enable imprecise
measure of object dimensions and location, thus improving system robustness
and adaptivity. In this section, we define dexterous grasp taxonomy
and implement grasping strategies on a robotic hand by extending our
prior work \cite{BRaoHHe-Self}.

\subsection{Grasp Definition}

\begin{figure}[h]
\includegraphics[width=0.8\columnwidth]{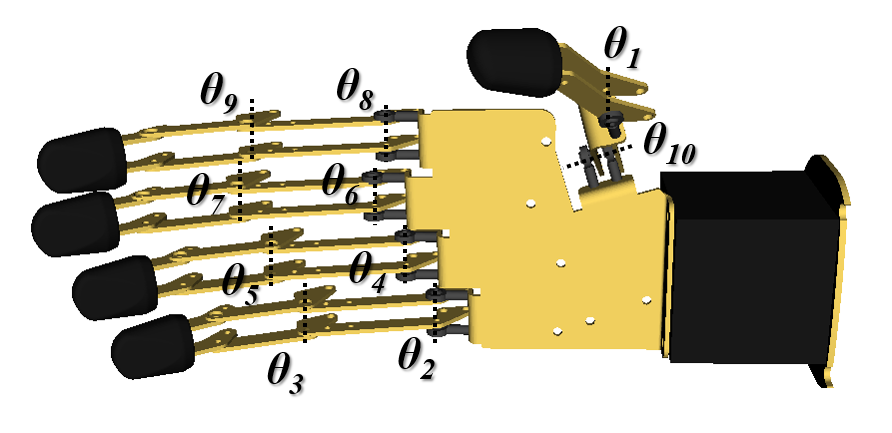}

\caption{AR10 Robotic Hand with 10 degrees of freedom (DoF).\label{fig:AR10-Robotic-Hand}}
\end{figure}

Grasps can be defined by the finger joint angles for a humanoid robotic
hand with multiple fingers. As shown in Fig. \ref{fig:AR10-Robotic-Hand},
the AR10 robotic hand used in this work has ten DoF and limited force
feedback. A grasp $G\in\mathbb{R}^{10}$ can therefore be defined
in the 10-dimensional joint configuration space as
\begin{equation}
G(t)=[\theta_{1}(t),\theta_{2}(t),\ldots\theta_{i}(t)]
\end{equation}
where each joint angle $\theta_{i}$ can continuously vary over the
operating range of the servos to result in infinite grasp patterns.
Owing to the redundant DoF, there may be multiple feasible grasp strategies
associated with one object, so it is intractable to design a deterministic
grasping model for various objects. We discretize the configuration
space and map configuration space into a space with reduced dimensionality
by
\begin{equation}
G=g(h,\boldsymbol{\alpha})
\end{equation}
where $h\in{h_{1},h_{2},..,h_{k}}$ represents human grasp topology
and $\mathbf{\alpha}$ is the scales that determine the completion
of the grasp. Each grasp topology $h_{k}=[\theta_{k1},\theta_{k2},...,\theta_{k10}]$
is a unique combination of joint angles representing one of the human
grasps with $\theta_{kj}$ chosen such that $h_{k}$ mimics a particular
human grasp type from the grasp taxonomy. The grasp topology $h$
spans the entire configuration space, and a grasp can be represent
by
\begin{equation}
G(t)=\boldsymbol{\alpha}(t).h
\end{equation}
whereby a range of grasps can be defined by human grasp topology $h$
and the time-variant completion scale $\boldsymbol{\alpha}(t)$. We
will learn a mapping between object features and grasp topology, and
implement the completion scale by inverse kinematics.

\subsection{Dexterous Grasp Taxonomy}

\begin{figure}[h]
\includegraphics[width=1\columnwidth]{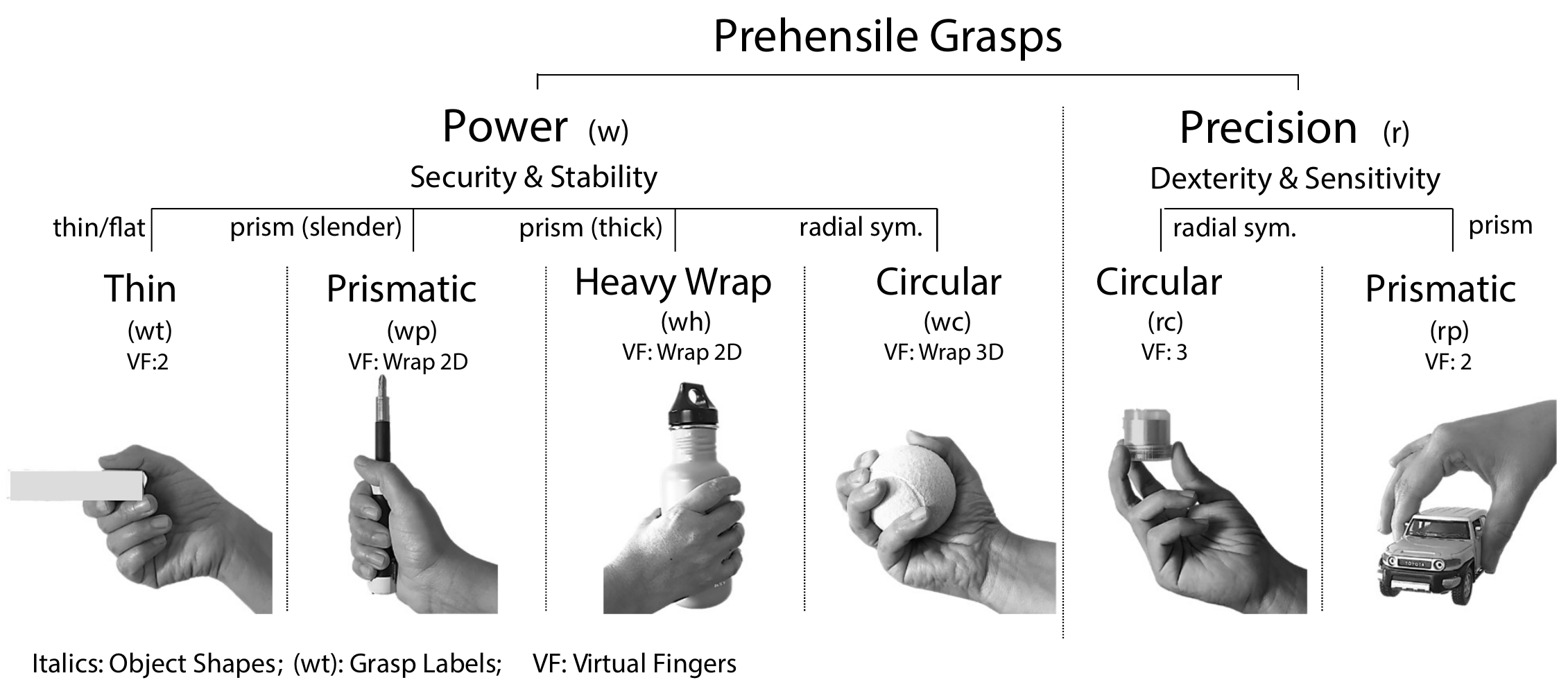}

\caption{Human Grasp Taxonomy derived from \cite{Cutkosky1989}\label{fig:Human-Grasp-Taxonomy}.}
\end{figure}

Robotic grasps can be made effective by emulating human grasps, and
we desire to choose a suitable taxonomy of human grasps. Although
comprehensive grasp taxonomy is available, we decided to adopt the
Grasp Taxonomy presented by Cutkosky \cite{Cutkosky1989} for this
work. Even within this taxonomy, we have restricted it to the six
higher level classifications, because the finer adjustments can be
obtained by combining these six grasp types with the scalar. The chosen
human grasp classification and the nomenclature for each grasp is
shown in Fig. \ref{fig:Human-Grasp-Taxonomy}, where the prefixes
$\lyxmathsym{\textquoteleft}w\lyxmathsym{\textquoteright}$ and $\lyxmathsym{\textquoteleft}r\lyxmathsym{\textquoteright}$
stand for Power and Precision grasps \cite{Cutkosky1989,NapierPrehenileMovements}.

\begin{figure}[h]
\includegraphics[width=0.98\columnwidth]{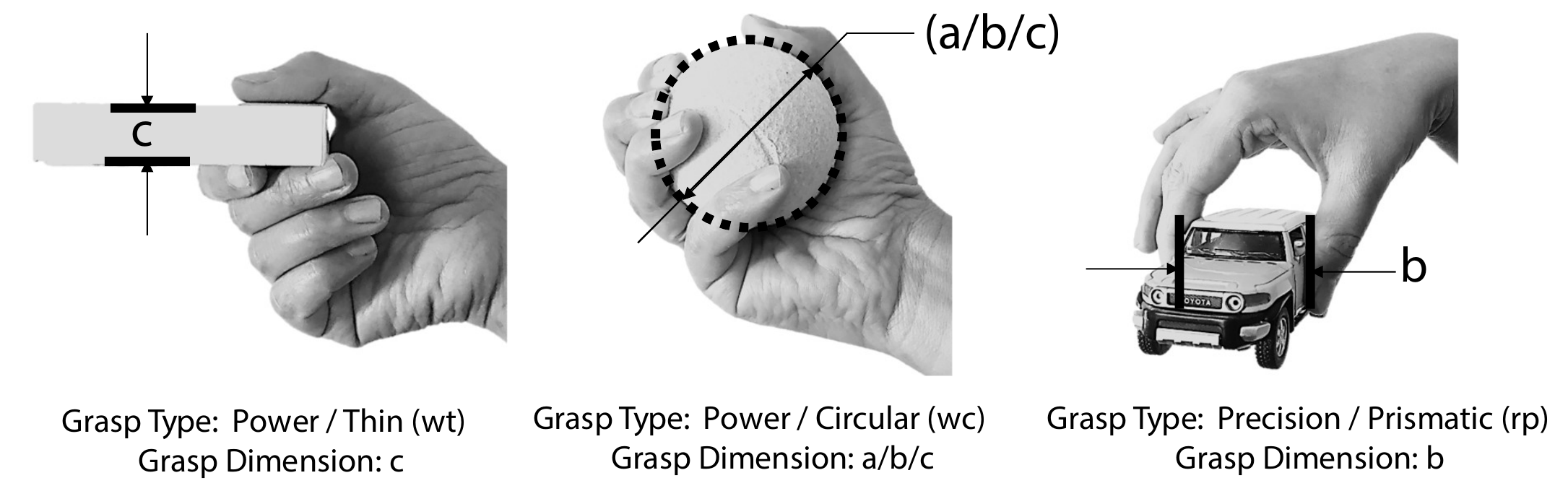}

\caption{Illustrations of grasp topology and size\label{fig:Illustrations-of-grasp}.}
\end{figure}

The grasp classification $h$ is one of the grasp types drawn from
the set of human grasp primitives 
\begin{equation}
h\in\{wt,wp,wh,wc,rp,rc\}
\end{equation}
Grasp scales are determined by the dimensions $a$, $b$, $c$ around
which the grasp closure occurs, as illustrated in Fig.~\ref{fig:Illustrations-of-grasp}.
This labeling convention is commonly adopted in the literature \cite{Feix2016l},
facilitating the computation of hand closure in forward and inverse
kinematics. Grasp dimension $d$ is defined as
\begin{equation}
d\in\{a,b,c,ab,bc,ac,abc\}
\end{equation}
The grasp scale is therefore a function of the grasp type and grasp
dimensions
\begin{equation}
\boldsymbol{\alpha}=f(h,d)\label{eq:2}
\end{equation}
For most object-grasp associations, we found the selection of grasp
type $h$ defines the selection of grasp dimensions and sizes; but
for certain object-grasp associations, the choices of grasp dimensions
were inconsistent for different attempts. Such confusion was mostly
observed for objects when $a/b\approx1$ or $b/c\approx1$. To address
this confusion problem, a new grasp classification was created by
concatenating the grasp type and dimension. For example, other than
\textquoteleft circular\textquoteright{} grasps, no other grasp type
uses \textquoteleft $abc$\textquoteright{} dimension; \textquoteleft thin\textquoteright{}
grasp cannot be executed along the longest dimension \textquoteright $a$\textquoteright .
The extended grasp taxonomy is defined as
\begin{equation}
H\in\{rc.ab,rc.bc,rp.b,rp.c,wc.abc,wh.bc,wh.c,wp.bc,wt.c\}\label{eq:grasp-topology}
\end{equation}

\subsection{Learning Grasping Strategies from Human Knowledge\label{subsec:Learning-Grasping-Strategies}}

Most studies \cite{NapierPrehenileMovements,Cutkosky1989,NatureMechanoreceptors}
attempting to understand and codify human grasps have concluded that
human grasp choice is a function of object affordances (geometry,
texture etc.) and the task requirements (forces, mobility, etc.).
Attempts to assign one most suitable grasp for a given object-task
combination have not been conclusive. The major problem is that even
for one specific object-task combination, there are multiple grasp
choices possible, which ofter appear to be arbitrary and not amenable
for deterministic modeling. Human grasp choices nevertheless do tend
to cluster when studied over a large set of objects. Both the clustering
effect and the confusion between grasp types can be seen in the data
presented by \cite{Feix2016l}, which shows that a single object could
be held in multiple different grasp types in the course of picking
or handling. There is no one-to-one mapping of one object to one grasp
type.

\begin{figure}[h]
\includegraphics[viewport=0bp 0bp 650bp 302bp,width=0.98\columnwidth]{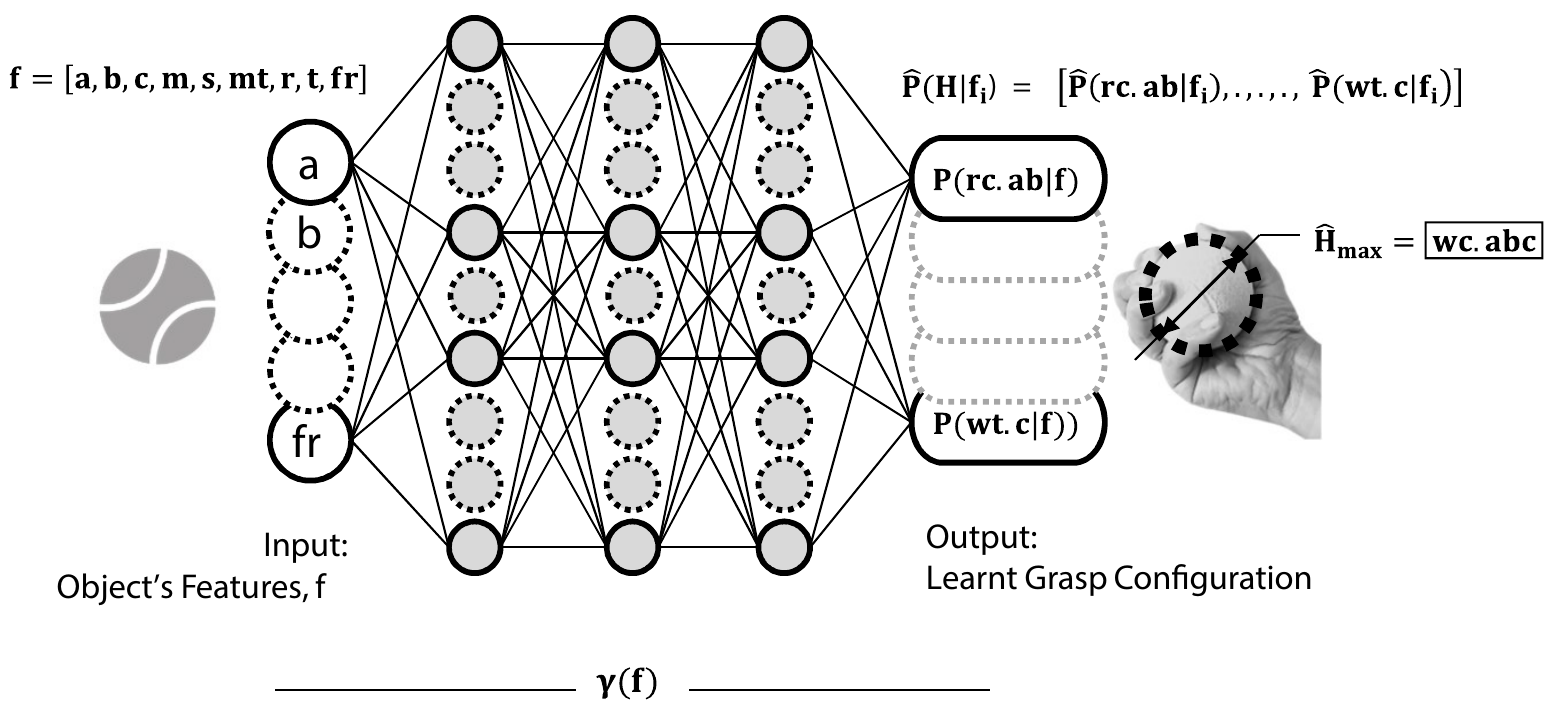}

\caption{Grasp Selection Neural Network Classification Model\label{fig:Grasp-Selection-Neural}.}
\end{figure}

The problem of grasp selection is therefore not selecting one ideal
grasp type but one of the many feasible grasp types in human grasp
taxonomy for the given context. To that end, we plan to learn the
mapping from features $f$ into grasp topology distributions
\begin{equation}
f\rightarrow P(H|f)=[P(rc.ab|f),P(rc.bc|f),\cdots]
\end{equation}
We designed a neural network to model the probability distribution
over all grasp classes $\hat{P}(H|f)$, as illustrated in Fig.~\ref{fig:Grasp-Selection-Neural}.
The network is designed with cross-entropy loss and optimized using
stochastic gradient decent algorithms. The loss function is defined
by cross entropy that measures the deviation between the ground truth
and predicted probability distribution
\begin{equation}
L(P(H|f),\hat{P}(H|f))=\sum_{i}\sum_{j}P(H_{j}|f_{i})\log\hat{P}(H_{j}|f_{i})
\end{equation}
where $i\in[1,N]$ with $N$ as the number of observations and $j$
is the index of grasp topology. 

The grasp with the maximum probability is chosen by
\begin{equation}
\hat{H}_{\max}^{*}=\arg\max_{j}\hat{P}(H|f^{*})
\end{equation}
where $f^{*}$ is the acquired object features. The predicted grasp
configuration $\hat{H}_{\max}^{*}$ contains information regarding
the grasp type and object dimension along which the grasp can be executed,
so $\hat{H^{*}}_{max}$ can be easily decomposed into grasp type $h^{*}$
and grasp dimension $d^{*}$, which can be used subsequently to calculate
robot hand configuration. The optimal grasp type is chosen as the
one corresponding to the highest probability from the predicted probability
distribution. 

Because the model predicts probability distributions, we defined two
scoring metrics for training and evaluation of the model. The predicted
grasp choice is scored as a success if the same grasp type was chosen
at least once in the human-knowledge database. The feasibility of
the grasp is scored as
\begin{equation}
F_{l}(P(H),\hat{P}(H))=\begin{cases}
1 & P(\hat{H}_{\max})>0\\
0 & P(\hat{H}_{\max})=0
\end{cases}\label{eq:Fl Feasibility Score}
\end{equation}
where 
\begin{equation}
\hat{H}_{\max}=\arg\max_{j}\hat{P}(H_{j}|f)
\end{equation}
is the grasp topology with the maximal probability, and $H$ is defined
in \eqref{eq:grasp-topology}. The feasibility score $F_{l}$ is representative
of the ability of the algorithm to pick a feasible grasp for a given
object. The match score metric $F_{m}$ is defined as
\begin{equation}
F_{m}(P(H),\hat{P}(H))=\begin{cases}
1 & P(\hat{H}_{\max})=P(H_{\max})\\
0 & P(\hat{H}_{\max})\neq P(H_{\max})
\end{cases}\label{eq:Fm Match Score}
\end{equation}
This match score is representative of the ability of algorithm to
predict the most frequently applied human grasp as the grasp with
the highest probability for a given object. In other words, $F_{m}$
is akin to the accuracy, had this grasp learning problem been treated
as a multi-class single label classification problem. This metric
is much more stringent and therefore we can expect the match score
$F_{m}$ to be always lower than feasibility score $F_{l}$
\begin{equation}
F_{m}(P(H),\hat{P}(H))\,\le\,F_{l}(P(H),\hat{P}(H))
\end{equation}
We used the feasibility score as the primary scoring metric, for the
objective is to find one feasible grasp that can be successfully executed
by a robot.

\subsection{Deploying Grasp Strategies\label{subsec:Action-=002013-Translating}}

Grasp types are determined by grasp topology, and the grasp size for
a particular grasp type corresponds to object dimensions. The grasp
size $d_{vf}$ is essentially the distance between the virtual fingers
of a particular grasp type $h^{*}$ \cite{Cutkosky1989}. The grasp
size $d_{vf}$ can be computed or estimated from object dimensions
based on geometric relations, as illustrated in Fig.~\ref{fig:Illustrations-of-grasp}.
The grasp types and sizes are implemented by inverse kinematics of
the hand and fingers.

We developed a multi-constrained inverse kinematics of the robotic
hand to deploy grasp topology and finger closure \cite{yamane2003natural}.
The multi-constrained inverse kinematics enables multi-point planning
of each finger in the process of hand closure and grasping. We considered
two levels of kinematic constraints in grasp strategy implementation:
high-priority and low-priority constraints. The distance between virtual
fingers (finger tip closure) meets  high-priority constraints on distal
phalanges, and the trajectories of the middle and proximal phalanges
satisfy low-priority constraints. The inverse kinematics transforms
the trajectory of points on fingers to angular joint velocities. 

\begin{figure}[H]
\includegraphics[width=0.98\columnwidth]{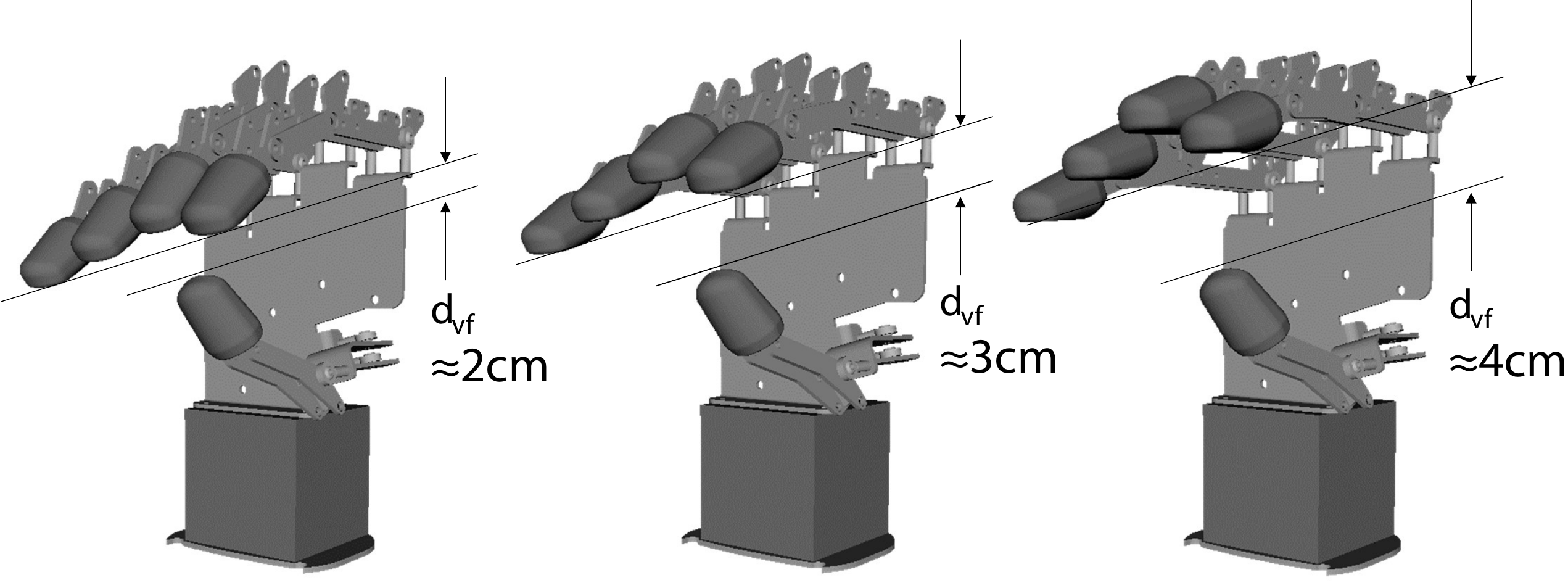}

\caption{Illustration of variations in virtual finger distance for the same
grasp type: prismatic grasp (rp).\label{fig:Illustration-of-variations}}
\end{figure}

The trajectory of $N$ finger tips (distal phalanges) $\boldsymbol{\alpha}(t):\left[0,T\right]\rightarrow\left[0,1\right]{}^{N}$
is planned for $d_{vf}$, which corresponds to the object dimensions.
Not all objects have regular geometric shapes and the calculated $d_{vf}$
may deviate from the actual size on such objects, as shown in Fig.~\ref{fig:Illustration-of-variations}.
We adopted a straight-line path 
\begin{equation}
d(\boldsymbol{\alpha})=d_{\textrm{start}}+\boldsymbol{\alpha}_{i}(d_{\mathnormal{vf}}-d_{\textrm{start}})
\end{equation}
where $\boldsymbol{\alpha}_{i}$ is the time scaling for the i\textsuperscript{th}
finger, and $d_{\textrm{start}}$ is the starting distance between
the virtual fingers. The closure of the fingers are controlled following
$\boldsymbol{\alpha}(t)$. The AR10 Robotic hand has 10 DOF with 9
actuators controlling the fingers and one controlling the opposing
thumb action. The robotic hand has limited and inaccurate force measurement
by the Force Sensitive Resistors (FSR) that are attached to each finger.
The closing of the fingers stops until sufficient contact forces are
measured. 

\begin{figure}[H]
\includegraphics[width=1\columnwidth]{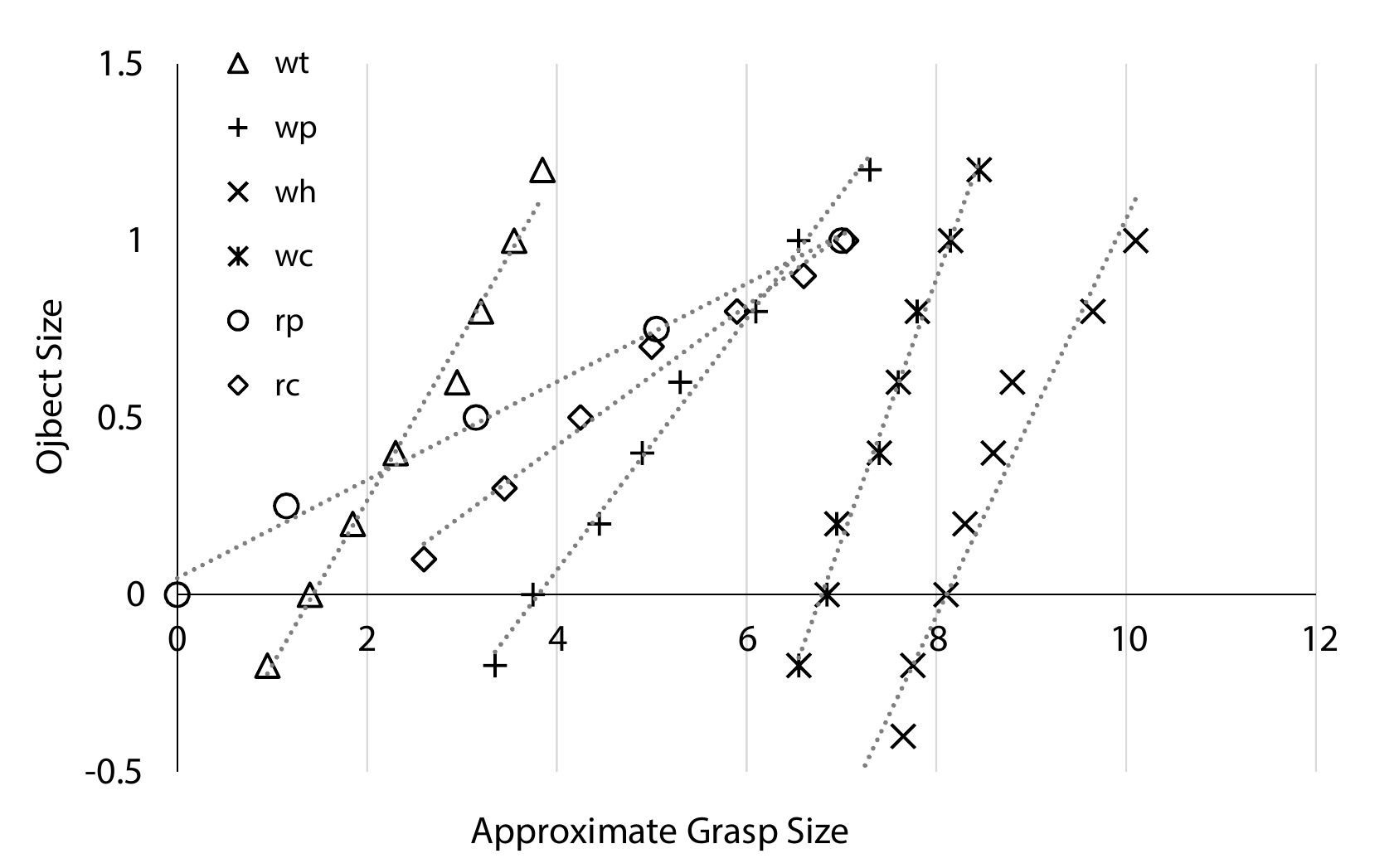}

\caption{The relation between virtual finger distances and object dimensions.\label{fig:The-relation-between}}
\end{figure}

In the experiments shown in Fig.~\ref{fig:The-relation-between},
we found the distance between virtual fingers $d_{vf}$ can be linearly
approximated by object dimensions $d_{o}$ within a limited range
of motion
\begin{equation}
d_{vf}=w_{1}d_{o}+w_{0}
\end{equation}
We learn the parameters $w_{1}$ and $w_{0}$ by generating data on
the physical robotic arm and fitting a linear model between $d_{o}$
and $d_{vf}$. For each grasp type, we varied the joint angles, measured
the distance between virtual fingers $d_{vf}$, and fit a linear regression
model to learn the parameters by minimizing least square errors. 

\section{Experiments \label{sec:Experimental-Analysis}}

We conducted the experiments to evaluate the performance of object
affordance acquisition, grasp strategy selection, and robot grasping.
Leveraging existing grasping database \cite{bullock2015yale}, we
collected more data for experiment and training, including object
attributes, short natural-language descriptions, and human grasp strategies.
The developed methods include natural-language parsing algorithms,
models for grasping strategy learning, object identification, and
robot grasping. We validate the methods and quantify the performance
on collected data by executing the grasps physically on a robot with
unfamiliar objects. 

\subsection{Experiment Setup}

\begin{figure}[h]
\begin{centering}
\includegraphics[width=0.98\columnwidth]{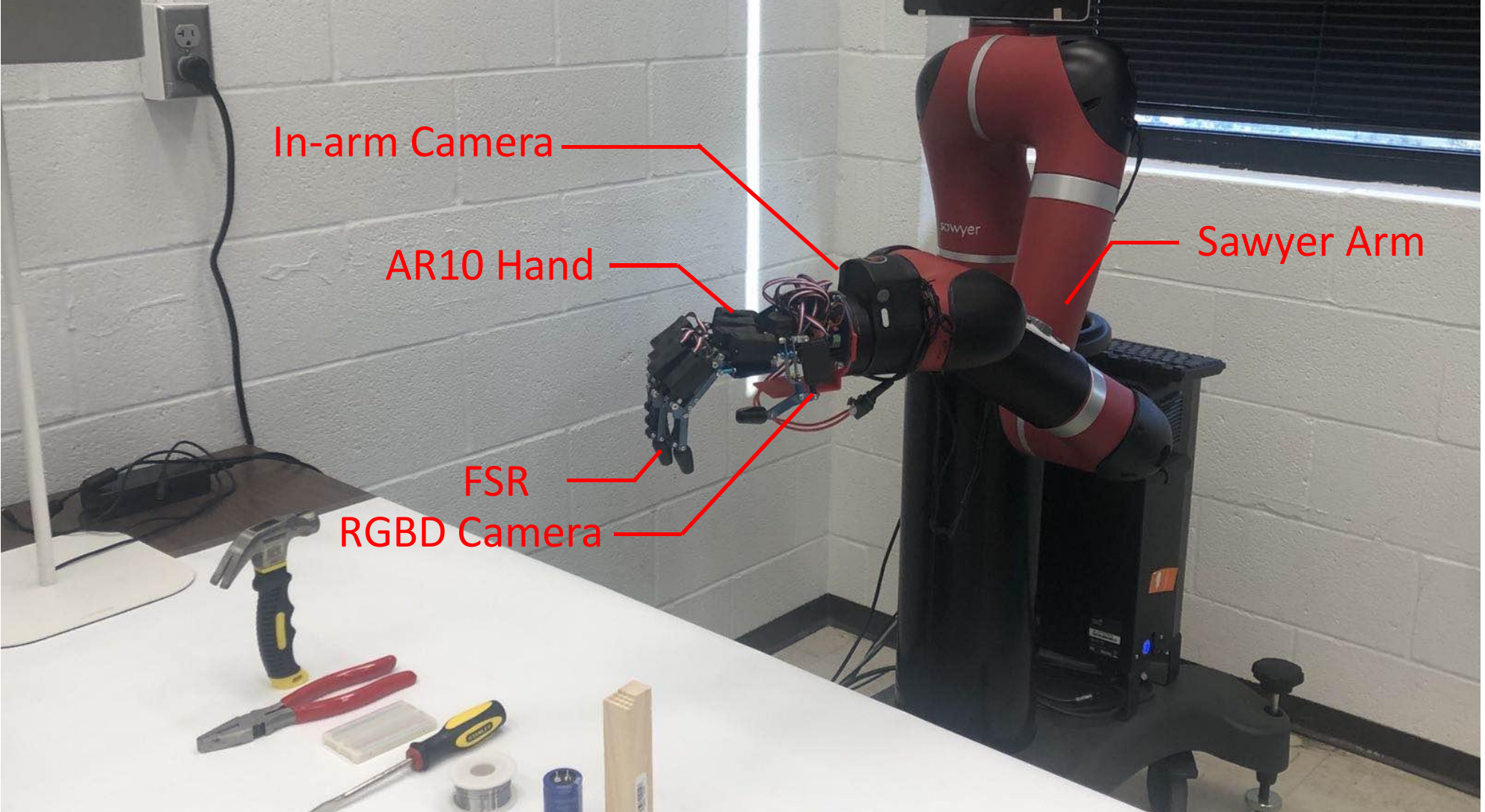}
\par\end{centering}
\caption{Robot Setup: Sawyer robotic arm with an AR10 robotic hand and in-hand
sensors. \label{fig:Robot-Setup:-Sawyer} }
\end{figure}

The anthropomorphic robotic hand integrates an AR10 robotic hand,
a Rethink Sawyer robot (with an in-arm camera), and an Intel RealSense
RGB-D camera installed on the wrist, as illustrated in Fig.~\ref{fig:Robot-Setup:-Sawyer}.
The robotic hand has limited force sensing through the Force Sensing
Resistors (FSR) attached to finger tips, so we focused on hand configurations
and planning and used the force sensors to examine contact conditions.
The goal of the experiment was to grasp an object, which was described
by a label or short incomplete description, in a proper manner to
the object affordance. We seamlessly integrated the models and modules
in ROS Python libraries that control the AR10 and Sawyer robots. The
experiment is set up as follows:
\begin{enumerate}
\item To emulate industrial scenarios with incomplete sensing, we only used
the camera to locate and identify an object. The objects or tools
for experiment are placed on the table with a fixed initial pose.
We utilized the localization and recognition algorithms developed
in our prior work \cite{fujian19robotic,fujian20common,yan2018semantics}.
The coordinates of the object on the table is back projected to the
robot frame.
\item We retrieved object affordances based on optional object descriptions,
e.g., ``A scientific calculator with plastic body. It is about fifteen
and half centimeters long, 8 centimeters wide and appears to be more
than one and half centimeters thick.'' The recognized object labels
are adopted when an object description is unavailable.
\end{enumerate}

\subsection{Data Collection}

\begin{figure}[h]
\includegraphics[width=0.98\columnwidth]{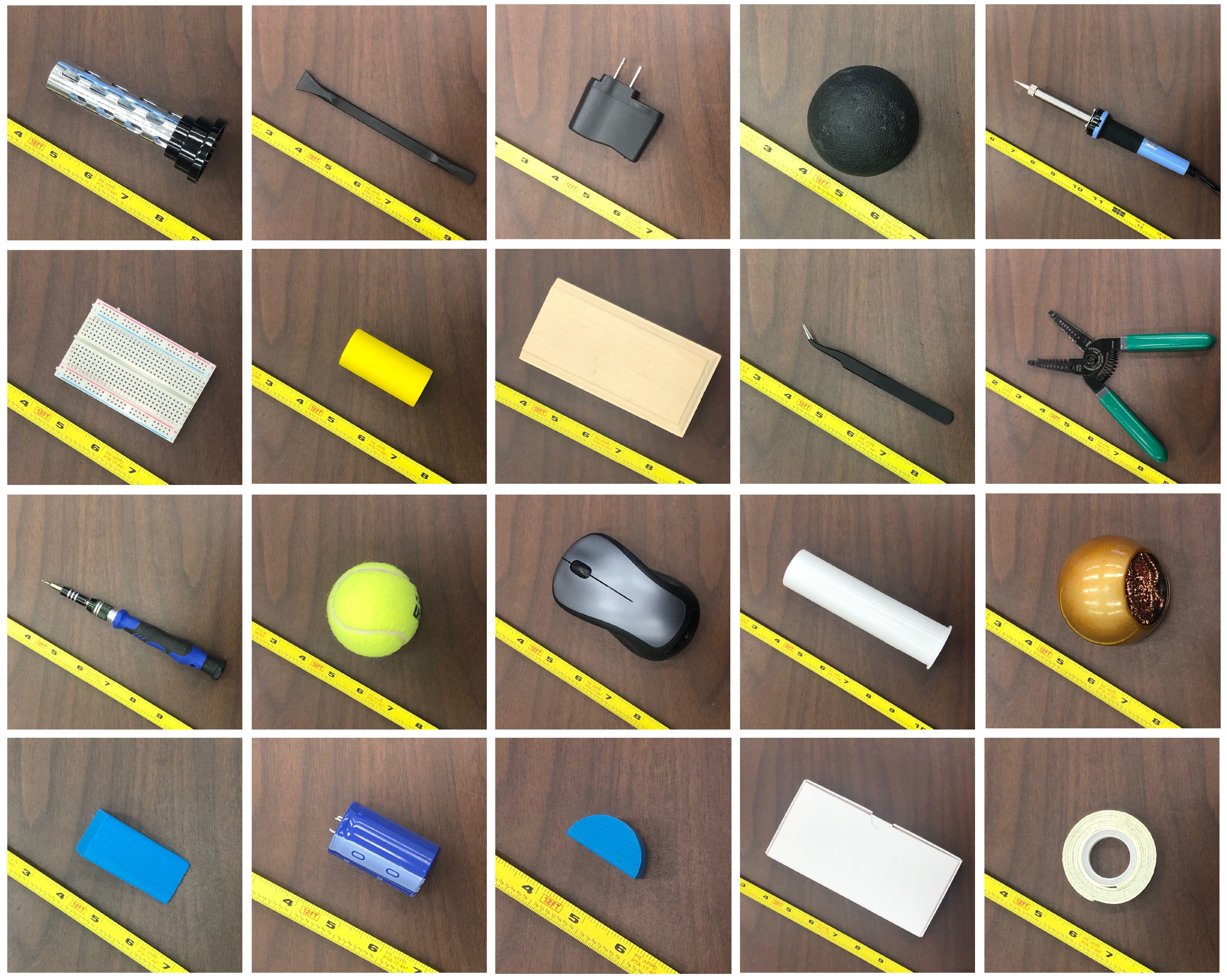}

\caption{Sample objects for experiments.\label{fig:A-random-sample}}
\end{figure}

We followed the design of the human grasping database \cite{bullock2015yale}
and extended the database by collecting grasping samples. Given the
ergonomics of human grasps, objects were selected to capture sufficient
variations in objects features and grasping strategies (samples are
shown in Fig.~\ref{fig:A-random-sample}). Short descriptions of
the objects are provided along with object labels, covering shape,
dimension, mass, rigidity, and texture if available. 

Grasping strategies depend on object affordances and tasks. In this
paper, the task definition is restricted to securely holding and supporting
an object in midair. The intent (task) of the grasp and the object\textquoteright s
position and orientation may influence the grasp strategies; however,
we kept these variables constant but focused on object features during
the experiment. There could be multiple ways of grasping for most
objects, so the experiment was designed to replicate potential grasping
multiple times. At each attempt, the subjects were encouraged to try
alternate ways of grasping the object while ensuring the comfort and
security of the grasp. At the end of this experiments, we had a frequency
distribution of human preferred grasp types. Sample data is shown
in Table \ref{tab:Grasp-(Frequency)-Labels}. It is inevitable that
the optimal grasp labels are subjective to some extent due to personal
preferences and background. There are indeed multiple feasible or
optimal strategies for one scenario. 

\begin{table}[h]
\caption{Grasp (Frequency) Labels -- Sample Of 10 Labels.\label{tab:Grasp-(Frequency)-Labels}}

\begin{tabular}{lllllllllll}
\toprule 
\textbf{\#} &
\textbf{Object} &
\begin{turn}{90}
\textbf{rc.ab}
\end{turn} &
\begin{turn}{90}
\textbf{rc.bc}
\end{turn} &
\begin{turn}{90}
\textbf{rp.b}
\end{turn} &
\begin{turn}{90}
\textbf{rp.c}
\end{turn} &
\begin{turn}{90}
\textbf{wc.abc}
\end{turn} &
\begin{turn}{90}
\textbf{wh.bc}
\end{turn} &
\begin{turn}{90}
\textbf{wh.c}
\end{turn} &
\begin{turn}{90}
\textbf{wp.bc}
\end{turn} &
\begin{turn}{90}
\textbf{wt.c}
\end{turn}\tabularnewline
\midrule
\addlinespace[2pt]
\rowcolor{lightgray}1 &
calculator &
0 &
0 &
5 &
2 &
0 &
0 &
0 &
0 &
2\tabularnewline
\addlinespace[2pt]
2 &
water bottle &
0 &
1 &
1 &
2 &
0 &
5 &
0 &
0 &
0\tabularnewline
\addlinespace[2pt]
\rowcolor{lightgray}3 &
wood cylinder &
0 &
2 &
5 &
2 &
0 &
0 &
0 &
0 &
0\tabularnewline
\addlinespace[2pt]
4 &
cardboard box &
0 &
0 &
5 &
2 &
0 &
0 &
0 &
0 &
2\tabularnewline
\addlinespace[2pt]
\rowcolor{lightgray}5 &
mini rubik's cube &
1 &
4 &
3 &
1 &
0 &
0 &
0 &
0 &
0\tabularnewline
\addlinespace[2pt]
6 &
wood wedge &
0 &
2 &
4 &
2 &
0 &
0 &
0 &
0 &
1\tabularnewline
\addlinespace[2pt]
\rowcolor{lightgray}7 &
wood disk &
4 &
0 &
2 &
2 &
0 &
0 &
0 &
0 &
1\tabularnewline
\addlinespace[2pt]
8 &
tennis ball &
1 &
1 &
2 &
1 &
4 &
0 &
0 &
0 &
0\tabularnewline
\addlinespace[2pt]
\rowcolor{lightgray}9 &
wood piece &
2 &
4 &
2 &
1 &
0 &
0 &
0 &
0 &
0\tabularnewline
\addlinespace[2pt]
10 &
plastic cap &
5 &
0 &
2 &
1 &
0 &
0 &
0 &
0 &
1\tabularnewline
\bottomrule
\addlinespace[2pt]
\end{tabular}
\end{table}

\subsection{Object Affordance Acquisition}

\begin{figure}[h]
\includegraphics[clip,width=1\columnwidth]{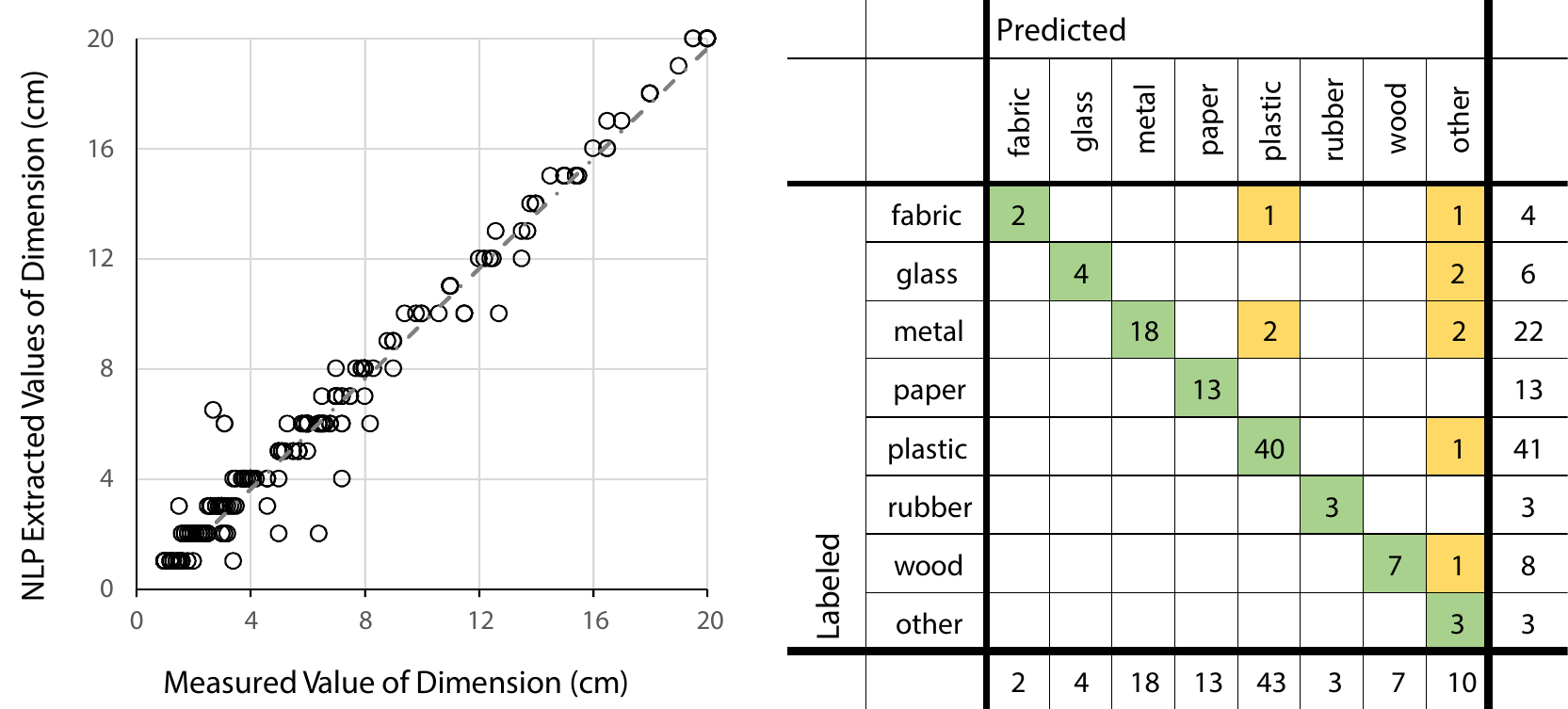}

\caption{Left: Parsed vs measured dimensions. Right: \textquoteleft Material\textquoteright{}
type prediction score.\label{fig:Left:-Parsed-vs}}
\end{figure}

The effectiveness of the natural-language parses was scored and validated
by the Ordinary Least Squares Regression model. The final model was
able to fit with an $R^{2}$ of $0.98$ overall for all dimensions
and $0.87$ for mass estimations. The regression fit for dimension
estimations is shown in Fig. \ref{fig:Left:-Parsed-vs}. The primary
source of errors in this step are the approximated dimensions in descriptions.
This results in larger percentage deviations when describing smaller
dimensions. Categorical labels for material, shape and rigidity were
also scored, where the scoring matrix for material classification
is shown in Fig.~\ref{fig:Left:-Parsed-vs}. The objects were classified
under ``other'' when description details are insufficient. The main
source of errors in material recognition is omitted features from
the description. This behavior stems from preconceived assumptions
that such features are very obvious and do not require specific mention.
For example, when the object is made of plastic, the subjects tend
to omit any mentions of the object\textquoteright s stiffness, assuming
plastic objects are rigid. 

Object affordances, including shapes, sizes, weights, and textures,
lead to specific grasping strategy \cite{NapierPrehenileMovements}.
We desired to prioritize the influence of the features on grasping
strategies. Such a prioritization would help the learning of grasping
strategies and the design of perception algorithms. To that end, we
used a recursive feature elimination (RFE) method along with a Random
Forest classifier to rank the features \cite{scikit-learn}. Starting
from the most significant factor, the ranking is dimension, shape,
mass, texture, material, stiffness, and fragility. The result is generally
in line with prior studies. It was interesting to note that the most
important dimension was the intermediate dimension $b$ followed by
the shortest dimension $c$ and then by the longest dimension $a$.
For most objects, especially larger ones, humans tend to grasp it
along the shorter dimension primarily because of comfort of holding. 

\begin{figure}[h]
\includegraphics[clip,width=1\columnwidth]{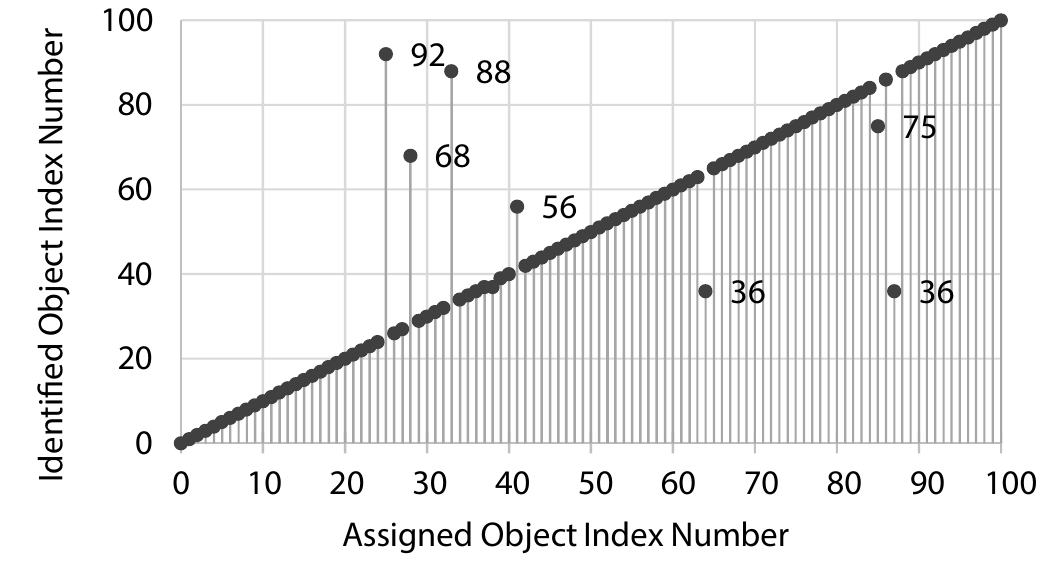}\caption{Plot of assigned object indices vs indices of objects chosen by the
Joint Probability Distance Metric. Points offset from line show mis-identified
objects. E.g. Object number 87 (a thin laptop) is mis-identified as
object 98 (a photo frame) \label{fig:Actual-vs.-Labeled}}
\end{figure}

\begin{figure*}
\begin{centering}
\includegraphics[width=1.6\columnwidth]{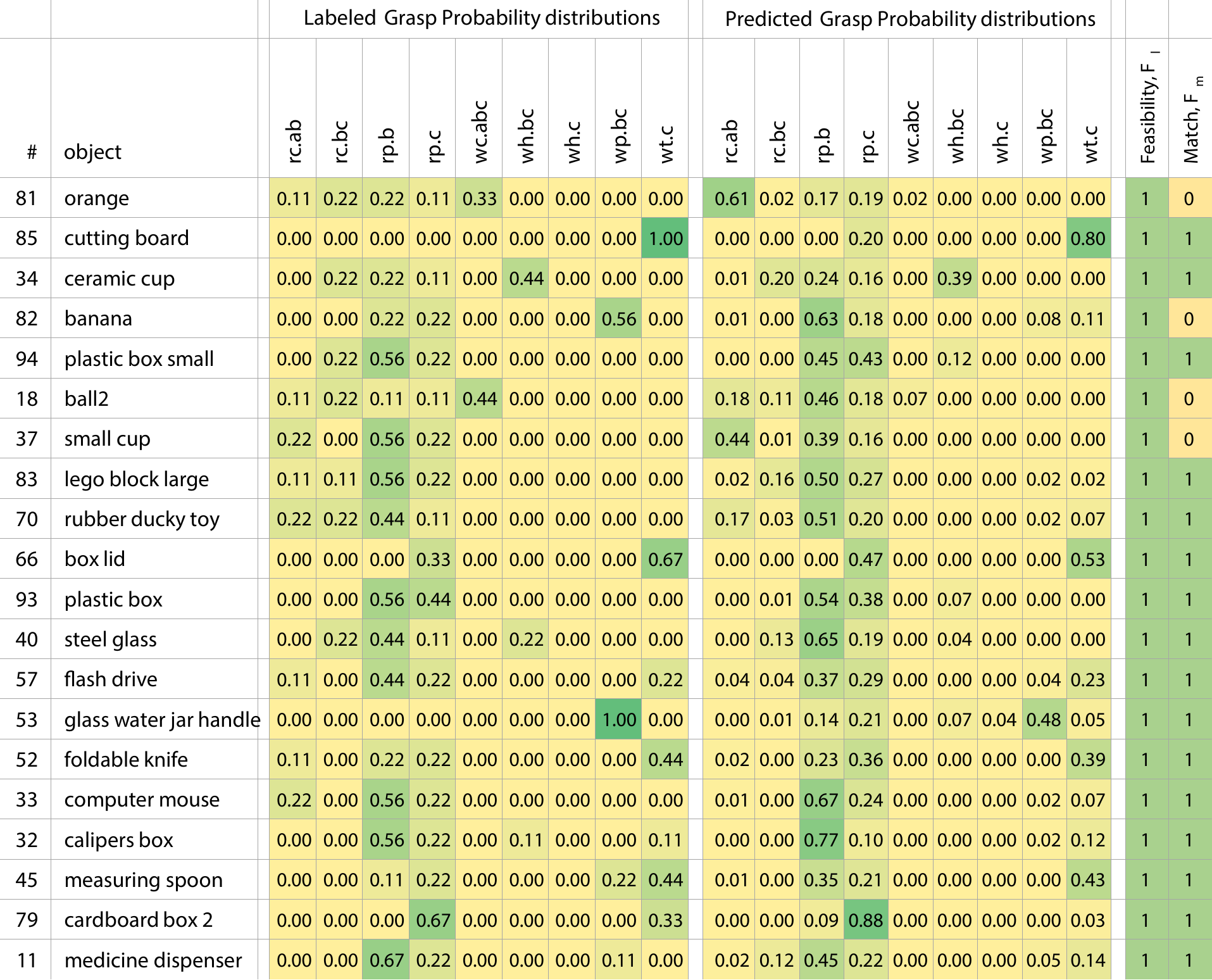}
\par\end{centering}
\caption{Sample grasping strategy determination: ground-truth vs. predicted
grasping.\label{tab:Probability-Distributions-Of}}
\end{figure*}

Object features parsed from object descriptions are used to acquire
full features of similar objects from the collected object database.
We performed contextual search based on the proposed distance metric,
and validate the performance of the methodology. The parsed object
features are commonly incomplete and inaccurate because of approximation
errors, missing information, and parsing errors. We used the subset
features to query and match the object being described to a similar
object in the database. The test results of acquisition of 100 random
objects are plotted in Fig.~\ref{fig:Actual-vs.-Labeled}. As the
confusion matrix show, the overall accuracy was 92\%. The errors primarily
stem from incomplete or insufficient information, similarity to multiple
objects, and wrong descriptions. The approach demonstrated robustness
in object acquisition under uncertainty, when partial object features
were incorrect. Despite the confusion, it was interesting to note
that the objects chosen by the algorithm were physically identical
to the objects being described. A mis-match is therefore not necessarily
detrimental to the grasp strategy. It is still possible to identify
the correct grasp with the wrong object as long as the object is physically
similar to the target.

\subsection{Robotic Grasping}

We developed the neural-network model to learn grasping strategies
as proposed in Sect.~\ref{subsec:Learning-Grasping-Strategies},
and optimized the model in terms of cross-entropy.  The input of
the model is the acquired object features, and the output is the grasping
strategies corresponding to human preference and knowledge. The grasping
strategies were represented by normalized probability distributions.
The results of grasping strategy determination with scores are reported
and compared to the ground truth in Fig.~\ref{tab:Probability-Distributions-Of}.
The feasibility score $F_{l}$ of the model is 100\%, which was defined
as the hit rate of the predicted grasp strategy in all human preferred
grasps. The experiment shows that the model's capability in picking
feasible (human validated) grasping. The match-score $F_{m}$, on
the other hand, measures the accuracy of the prediction considering
only the most preferred human grasping. The experiment demonstrated
that the max-match rate was around 80\% for the test objects. 

\begin{figure}
\includegraphics[width=1\columnwidth]{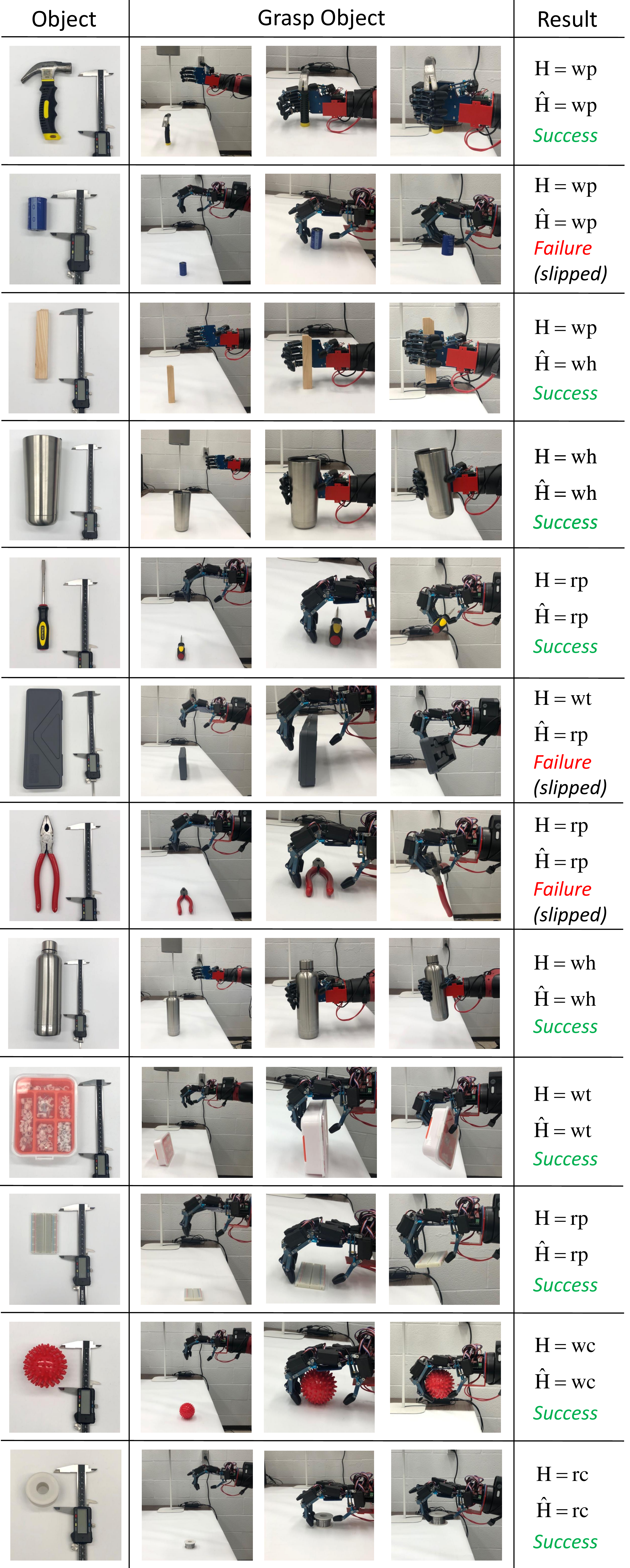}\caption{Robot grasp trials with 10 test objects. $H$ stands for human labeled
grasps; $\hat{H}$ stands for Learned grasp.$\hat{H}$ in red color
indicates mismatch with human grasp preference. \textquoteright Failure\textquoteright{}
indicates robot\textquoteright s inability to grasp the object. \textquoteleft Success\textquoteright{}
indicates that the grasp was successfully executed on the AR10 Robotic
Hand. More percentage of failure cases are demonstrated for better
understanding of the limitation of the method.\label{fig:Robot-grasp-trials}}
\end{figure}

To further examine grasping performance, we performed grasping experiments
on the robotic hand platform. Test objects are placed on the table
with a default initial orientation, and the objects are located and
identified by the developed object recognition algorithms. A short
description of each object was provided to cover its estimated dimension
and materials. The robot autonomously chose the grasp strategies according
to the acquired object affordance from the object database. The success
of a grasp is validated by the security of grasping after brief maneuvers
including lifting, holding and placing. A grasp is deemed as a success
if the object does not fall during the maneuvers. The overall success
rate of grasping was around 89\%, and some experiment results are
shown in Fig.~\ref{fig:Robot-grasp-trials}. One failure case was
the grasping of the capacitor, where the model predicted the most
preferred human grasping strategy but failed to securely grasp the
capacitor; the failure could be attributed to the limitation of hand
dexterity or inadequate friction. Another failure case was the grasping
of the plastic container, where the model predicted a different grasping
strategy (rp) instead of the most preferred human grasp (wt). Though
the predict strategy was one of the grasps used by humans and hence
valid, the robot could not manage to secure the object. The other
failure case was the grasping of the plier, where the model predicted
the most preferred human grasping strategy (rp), but the plier changed
formation during the grasping and fell. The developed system does
not possess the capability in adjusting strategies during the process
of grasping. There were many successful cases of grasping, where the
predicted grasping strategies were different from the most preferred
ones. 

Grasping involves a series of decision making processes using experience
and knowledge of the physical world as the basis. Human grasp strategies
vary even when most of contextual variables are fixed. The existing
human grasping data validated this interpretation: most objects are
associated with multiple grasping strategies for the same tasks. Furthermore,
it is still impractical to grasp one object without measuring or acquiring
contextual information. Familiarity with a specific object is important
knowledge that robots need to acquire before attempting a grasp with
the assistance of object recognition algorithms. In an event of confusion,
it still identifies the closest object with similar physical features,
ensuring that there is enough information to continue onto grasp execution.

We designed the sequential machine-learning model to emulate human
decision-making processes. In such a sequential model, errors from
one model could permeate or propagate into the next model. In the
experiment, while there were errors in each stage of and prediction,
we did not observe a significant impact on the final grasp execution
on the robot. The primary reason could be that, while human grasping
is complex, it is also highly resilient to external perturbations
of contextual variables. When modeling robot gasping using human grasp
primitives, this resilience behavior was emulated as well. For example,
there are multiple ways to grasp an object, so it is less likely to
choose a wrong grasp as we have seen from the results of our deep
learning model (100\% match score). The other reason is that the experiment
objects were designed with the intent of handling by five fingered
hands, so when there is a mis-calculation, e.g., grasp dimensions,
the fingers conform to the object shape and still result in a secure
grasp.

\section{Conclusion}

This paper has demonstrated the approach to applying a proper strategy
to grasp an object without complete sensing of object affordance.
The framework of grasping strategy determination, object affordance
acquisition, and robotic grasping deployment was developed through
a combination of probabilistic and machine learning models in order
to teach robots to grasp unfamiliar objects. The strategy determination
can predict human grasping knowledge with a 100\% feasibility score
and a 80\% match score. The designed distance metric outperformed
other popular distance metrics and achieved overall 92\% accuracy
in object feature acquisition. These learning models could be extended
to include additional inputs for position, orientation, and grasp
intent for a broader applicability. With more sophisticated robotic
arms built with multiple tactile sensors and precision force control,
we can train models with more data and realize better performance
in executing the grasps on real world objects. In summary, the experiments
show that emulating human behavior is a practical way to build an
autonomous robotic hand capable of adapting to unfamiliar environment. 

\bibliographystyle{unsrt-fr}
\bibliography{14_home_rilab3_Desktop_Bpaper_RoboticGraspfromNLDescription_RoboticGraspfromNLDescription}

\end{document}